\documentclass[conference]{IEEEtran}
\usepackage{times}

\usepackage[numbers]{natbib}
\usepackage{multicol}
\usepackage[bookmarks=true,draft]{hyperref}
\usepackage{algorithmic}
\usepackage[ruled,linesnumbered,vlined]{algorithm2e}
\usepackage{amsmath}

\usepackage[T1]{fontenc} % Use 8-bit encoding that has 256 glyphs
\usepackage{fourier} % Use the Adobe Utopia font for the document - comment this line to return to the LaTeX default
\usepackage[english]{babel} % English language/hyphenation
\usepackage{lipsum} % Used for inserting dummy 'Lorem ipsum' text into the template
\usepackage{amsmath,bm}

\usepackage{amssymb}
\usepackage{amsfonts}
\usepackage{algorithmic}
\usepackage[ruled,linesnumbered,vlined]{algorithm2e}
\usepackage[final]{graphicx}
\usepackage{textcomp}
\usepackage{xcolor}
\usepackage{etoolbox}
\usepackage{subcaption}
\usepackage{comment}
\usepackage{cleveref}
\usepackage[english]{babel}
\usepackage[utf8]{inputenc}
\usepackage{mathtools}
\usepackage{flushend}
\usepackage{wrapfig}

\usepackage{subcaption}

\begin{document}

\title{Error-Aware Policy Learning: Zero-Shot Generalization in Partially Observable Dynamic Environments}

% You will get a Paper-ID when submitting a pdf file to the conference system
%\author{Author Names Omitted for Anonymous Review. Paper-ID [234] }

\author{\authorblockN{Visak Kumar}
\authorblockA{School of Interactive Computing\\
Georgia Institute of Technology\\
Email: visak3@gatech.edu}
\and
\authorblockN{Sehoon Ha}
\authorblockA{School of Interactive Computing\\
Georgia Institute of Technology\\
Email: sehoonha@gatech.edu}
\and
\authorblockN{C. Karen Liu}
\authorblockA{Computer Science Department\\
Stanford University\\
Email: karenliu@cs.stanford.edu}

}
%\and
%\authorblockN{James Kirk\\ and Montgomery Scott}
%\authorblockA{Starfleet Academy\\
%San Francisco, California 96678-2391\\

% avoiding spaces at the end of the author lines is not a problem with
% conference papers because we don't use \thanks or \IEEEmembership

%Homer Simpson\authorrefmark{2},
%Montgomery Scott\authorrefmark{3} and
%Eldon Tyrell\authorrefmark{4}}
%Telephone: (800) 555--1212, Fax: (888) 555--1212}
%\authorblockA{\authorrefmark{4}Tyrell Inc., 123 Replicant Street, Los Angeles, California 90210--4321}}

\maketitle

\begin{abstract}
Simulation provides a safe and efficient way to generate useful data for learning complex robotic tasks. However, matching simulation and real-world dynamics can be quite challenging, especially for systems that have a large number of unobserved or unmeasurable parameters, which may lie in the robot dynamics itself or in the environment with which the robot interacts. We introduce a novel approach to tackle such a sim-to-real problem by developing policies capable of adapting to new environments, in a zero-shot manner. Key to our approach is an error-aware policy (EAP) that is explicitly made aware of the effect of unobservable factors during training. An EAP takes as input the predicted future state error in the target environment, which is provided by an error-prediction function, simultaneously trained with the EAP. We validate our approach on an assistive walking device trained to help the human user recover from external pushes. We show that a trained EAP for a hip-torque assistive device can be transferred to different human agents with unseen biomechanical characteristics. In addition, we show that our method can be applied to other standard RL control tasks.
\end{abstract}

\IEEEpeerreviewmaketitle
%% editing comment

%\newcommand{\cmt}[1]{\textcolor{red}{\textbf {#1}}}
\newcommand{\cmt}[1]{}
\newcommand{\visak}[1]{\textcolor{blue}{{Visak: #1}}}
\newcommand{\sehoon}[1]{\textcolor{orange}{{Sehoon: #1}}}
\newcommand{\revised}[1]{\textcolor{blue}{{#1}}}
\newcommand{\karen}[1]{\textcolor{red}{{Karen: #1}}}
\newcommand{\newtext}[1]{#1}
\newcommand{\original}[1]{\textcolor{magenta}{Original: #1}}
\newcommand{\eqnref}[1]{Equation~(\ref{eq:#1})}
\newcommand{\figref}[1]{Figure~\ref{fig:#1}}
\newcommand{\algref}[1]{Algorithm~\ref{alg:#1}}
\newcommand{\tabref}[1]{Table~\ref{tab:#1}}
\newcommand{\secref}[1]{Section~\ref{sec:#1}}

%% ignore text
\long\def\ignorethis#1{}

%% abbreviations
\newcommand{\etal}{{\em{et~al.}\ }}
\newcommand{\eg}{e.g.\ }
\newcommand{\ie}{i.e.\ }

%% reference shortcuts
\newcommand{\figtodo}[1]{\framebox[0.8\columnwidth]{\rule{0pt}{1in}#1}}

%\renewcommand{\eqref}[1]{Equation~(\ref{eq:#1})}

%% frequently used mathematical structures

\newcommand{\pdd}[3]{\ensuremath{\frac{\partial^2{#1}}{\partial{#2}\,\partial{#3}}}}

%% New commands for Sehoon!
\newcommand{\mat}[1]{\ensuremath{\mathbf{#1}}}
\newcommand{\set}[1]{\ensuremath{\mathcal{#1}}}

% math macros
\newcommand{\vc}[1]{\ensuremath{\mathbf{#1}}}
\newcommand{\vEndEff}{\ensuremath{\vc{d}}}
\newcommand{\vRelMove}{\ensuremath{\vc{r}}}
\newcommand{\sSet}{\ensuremath{S}}

\newcommand{\vControl}{\ensuremath{\vc{u}}}
\newcommand{\vPoint}{\ensuremath{\vc{p}}}
\newcommand{\sSpringCoef}{{\ensuremath{k_{s}}}}
\newcommand{\sDamperCoef}{{\ensuremath{k_{d}}}}
\newcommand{\vHandle}{\ensuremath{\vc{h}}}
\newcommand{\vForce}{\ensuremath{\vc{f}}}

\newcommand{\mTransChain}{\ensuremath{\vc{W}}}
\newcommand{\mRotateTrans}{\ensuremath{\vc{R}}}
\newcommand{\sJoint}{\ensuremath{q}}
\newcommand{\vJoint}{\ensuremath{\vc{q}}}
\newcommand{\mJoint}{\ensuremath{\vc{Q}}}
\newcommand{\mMass}{\ensuremath{\vc{M}}}
\newcommand{\sMass}{\ensuremath{{m}}}
\newcommand{\vGravity}{\ensuremath{\vc{g}}}
\newcommand{\vConstr}{\ensuremath{\vc{C}}}
\newcommand{\sConstr}{\ensuremath{C}}
\newcommand{\vCOM}{\ensuremath{\vc{x}}}
\newcommand{\sGeneralForce}[1]{\ensuremath{Q_{#1}}}
\newcommand{\vStateVar}{\ensuremath{\vc{y}}}
\newcommand{\vControlVar}{\ensuremath{\vc{u}}}
\newcommand{\tr}[1]{\ensuremath{\mathrm{tr}\left(#1\right)}}

%%%%%%%%%%%%%%%%%%%%%%%%%%%%%%%%%%%%%%%%%%%%%%%%%%%%%%%%%%%%%%%%%%%
%
% Here are a bunch of macros, mostly for math.
%
%%%%%%%%%%%%%%%%%%%%%%%%%%%%%%%%%%%%%%%%%%%%%%%%%%%%%%%%%%%%%%%%%%%

\renewcommand{\choose}[2]{\ensuremath{\left(\begin{array}{c} #1 \\ #2 \end{array} \right )}}

\newcommand{\gauss}[3]{\ensuremath{\mathcal{N}(#1 | #2 ; #3)}}

\newcommand{\pctab}{\hspace{0.2in}}
\newenvironment{pseudocode} {\begin{center} \begin{minipage}{\textwidth}
                             \normalsize \vspace{-2\baselineskip} \begin{tabbing}
                             \pctab \= \pctab \= \pctab \= \pctab \=
                             \pctab \= \pctab \= \pctab \= \pctab \= \\}
                            {\end{tabbing} \vspace{-2\baselineskip}
                             \end{minipage} \end{center}}
\newenvironment{items}      {\begin{list}{$\bullet$}
                              {\setlength{\partopsep}{\parskip}
                                \setlength{\parsep}{\parskip}
                                \setlength{\topsep}{0pt}
                                \setlength{\itemsep}{0pt}
                                \settowidth{\labelwidth}{$\bullet$}
                                \setlength{\labelsep}{1ex}
                                \setlength{\leftmargin}{\labelwidth}
                                \addtolength{\leftmargin}{\labelsep}
                                }
                              }
                            {\end{list}}
\newcommand{\newfun}[3]{\noindent\vspace{0pt}\fbox{\begin{minipage}{3.3truein}\vspace{#1}~ {#3}~\vspace{12pt}\end{minipage}}\vspace{#2}}

\newcommand{\key}{\textbf}
\newcommand{\fun}{\textsc}

%\def\shortcite{\def\citename##1{}\@internalcite}

% Local Variables:
% TeX-master: "paper"
% End:

\section{INTRODUCTION}
%\karen{The first two paragraphs can be replaced by one paragraph talking about the challenges of sim-to-real transfer across highly unobservable dynamics. Unobservable dynamics can be due to the complexity in agent mechanisms or due to the complexity in the environment the agent interacts with. For example, a complex quadruped has many unobserved parameters that cannot be reliably identified. Or a wearable robot that has to interact with a highly unobserved environment, the human. You can go into detail to describe these two examples a bit. The second paragraph will be mainly the current fourth paragraph. After you introduce EAP, the next paragraph is about how you evaluate EAP. You can add more emphasis on assistive example here by saying that we will evaluate EAP on the wearable device as it is our primary motivation for developing EAP. The text in the current third paragraph can be reused here, as well as the current fifth paragraph. At the end, you can mention that we also test EAP on a quadruped and other benchmark problems. Depending on how much hardware progress you make, we can come back to add more emphasis on the quadruped.}

Simulation has a growing role in learning-based methods to design control policies for robots as it provides a safe and efficient way to generate useful data. However, for robotic agents that are governed by complex dynamics or interacting with a complex environment, it is challenging to identify a model that captures the real-world dynamics accurately, giving rise to the so called sim-to-real problem in transferring control policies. Among many factors responsible for the sim-to-real gap, we are interested in addressing the challenges involving unobserved or unmeasurable model parameters, which may lie in the robot dynamics itself or in the environment with which the robot interacts. For example, for assistive robots that aid in locomotion, the dynamics of the wearable robot is closely coupled with humans, who show remarkably large variations in their movements, such as variations in muscle activation dynamics, muscle maximum force, fatigue, robot-human interaction dynamics, delay in sensing/actuation, all of which are difficult to measure and parameterize accurately. Similarly, other types of robots, such as quadrupeds and bipeds, also require accurate complex contact dynamics, contact parameters, motor dynamics, delay in the system, difficult to measure or model correctly.

Two broad approaches have been proposed to address the sim-to-real issue: 1) Domain randomization and domain adaptation methods \cite{openai2018learning,TobinFRSZA17,PintoDSG17,Pengdr,rajeswaran2017epopt} which aim to learn robust or universal policies by training them with variations in the model parameters. These methods require manual engineering of the range in which the parameters are varied. For highly complex systems and environments, domain randomization or adaptation often determines a subset of parameters to be observable and leaves the unpredictable effect of unobsered parameters to chance. 2) System identification methods \cite{pmlr-v100-xie20a,Jieetal,Hwangboeaau5872,chebotar2018closing}, aim to identify accurate models of the real robot to bring simulation closer to reality. System identification can be interleaved with policy learning---deploying the current policy in the target environment to collect more data to further improve the dynamic model \cite{yu2019simtoreal,peng2020data,belkhale2020modelbased}. Since the task-relevant training data is difficult to acquire from the real world, system identification also needs to determine a subset of parameters to be observable to avoid over fitting. Neither approach has demonstrated reliable ability to transfer policies to the real world when the combined dynamics of agent and environment is highly unobservable.

In this work, we introduce a novel approach to tackle sim-to-real problems in which the environment dynamics has high variance and is highly unobservable. While our approach is motivated by physical assistive robotic applications, the method can be applied to other tasks in which many dynamic parameters are challenging to model.  We propose to train a policy explicitly aware of the effect of unobservable factors during training, called an Error-Aware policy (EAP). Akin to the high-level idea of meta learning, we divide the dynamical environments into training and validation sets and "emulate" a reality gap in simulation. Instead of estimating the model parameters that give rise to the emulated reality gap, we train a function that predicts the deviation (i.e. error) of future states due to the emulated reality gaps. Conditioned on the error predictions, the error-aware policies (EAPs) can learn to overcome the reality gap, in addition to mastering the task. 

The main application in this work is to learn an error-aware policy for assistive device control, such as a hip-exoskeleton that helps the user to recover balance during locomotion. From biomechanical data of human gait, we model multiple virtual human walking agents, each varying in physical characteristics as well as parameters that affect the dynamics such as joint damping, torque limits, ground friction, and sensing and actuation delay. We then train a single policy on this group of human agents and show that that the learned EAP works effectively when tested on a different human agent without needing additional data. We extend the prior work, \cite{kumar2019learning}, that trained a control policy for push-recovery assistive device for just one simulated human agent, and develop an algorithm that enables the learned policy to transfer to other human agents with unseen biomechanical characteristics.

We evaluate our approach on assistive wearable device by quantifying the stability and gait characteristics generated by an unseen human agent wearing the device with the trained EAP. We present a comprehensive study of the benefits of our approach over prior zero-shot methods such as universal policy (UP) and domain randomization (DR). We also provide results on some standard RL environments, such as Cartpole, Hopper, 2D walker and a quadrupedal robot.

\section{Related work}

\subsection{Transfer of RL policies}

A popular approach to transfer control policies is Domain randomization (DR). DR methods   \cite{openai2018learning,TobinFRSZA17,PintoDSG17,Pengdr,rajeswaran2017epopt} propose to train policies that are robust to variations in the parameters that affect the system dynamics. Although some of these methods have been validated in the real world \cite{openai2018learning,Pengdr}, DR often requires manual engineering of the range in which the parameters are varied to make sure that the true system model lies within the range of variation. For a complex robotic system, it is often challenging to estimate the correct range of all the parameters because a large range of variation could lead to lower task performance, whereas a smaller range leads to less robust policies. To address the demanding sample budget issue with domain randomization, \cite{muratore2021dataefficient} presented a data-efficient domain randomization algorithm based on bayesian optimization. The algorithm presented in Mehta et al \cite{mehta2019active} actively adapts the randomization range of variation to alleviate the need for exhaustive manual engineering. Ramos et al \cite{ramos2019bayessim} proposed an approach to infer the distribution of the dynamical parameters and showed that policies trained with randomization within this distribution can transfer better. 

\begin{comment}
\cite{ramos2019bayessim} - bayesim
\cite{wang2020modular} - modular policy training for transfer
\cite{sonar2020invariant} - Invariant policy optimization
\cite{zhou2019environment} - environment probing policies
\cite{rajeswaran2017epopt} - epopt
\cite{yu2019learning} - strategy optimization
\cite{jegorova2020adversarial} -  
\cite{mehta2019active} - active domain randomization where current policy does poorly
\cite{desai2020stochastic} - grounded actions
\cite{allevato2020tunenet} - one shot residual transfer
\end{comment}

Careful identification of parameters using data from the real world, popularly known as system identification, has also shown promising results in real-world robots.
Tan et al \cite{Jieetal} and Hwangbo et al \cite{Hwangboeaau5872} carefully identified the actuator dynamics to bring the source environment closer to the target, Xie et al \cite{pmlr-v100-xie20a} also demonstrated that careful system identification techniques can transfer biped locomotion policies from simulation to real-world. Jegorova et al \cite{jegorova2020adversarial} presented a technique that improves on existing system identification techniques by borrowing ideas from generative adversarial networks (GAN) and showed improved ability to identify the parameters of a system. Similarly, Jiang et al \cite{jiang2021simgan} presented a SimGAN algorithm that identifies a hybrid physics simulator to match the simulated trajectories to the ones from the target domain to enable policy adaptation.
Yu et al \cite{yu2017preparing} developed a method that combines online system identification and universal policy to enable identifying dynamical parameters in an online fashion. Citing the difficulty in obtaining meaningful data for system identification, \cite{zhou2019environment} developed an algorithm that probes the target environment to provide more information about the dynamics of the environment. A few model based approaches have also been successful in transferring policies to a target domain \cite{song2020provably,desai2020stochastic,DataEffLeg}. 

Another popular approach of transferring policies includes utilizing data from the target domain to improve the policy. Chebotar et al \cite{chebotar2018closing} presented a method that interleaves policy learning and system identification, however this requires deploying the policy in the target domain every few iterations. This method would be impractical for a system that interacts closely with a human because of safety concerns. Yu et al \cite{yu2019simtoreal} and Peng et al \cite{RoboImitationPeng20} presented latent space adaptation techniques where the policy is adapted in the target domain by searching for a latent space input to the policy that enables successful transfer. Exarchos et al \cite{exarchos2020policy} also presented an algorithm that achieved policy transfer using only kinematic domain randomization combined with policy adaptation in the target domain, similar to \cite{yu2019simtoreal}. 

Yu et al~\cite{yu2020learning} proposed Meta Strategy Optimization, a meta-learning algorithm for training policies with latent variables that can quickly adapt to new scenarios with a handful of trials in the target environment.
Among the methods that use data from the target domain also include meta-learning approaches like  Bhelkale et al \cite{belkhale2020modelbased}, in which a model-based meta-reinforcement learning algorithm was presented to account for changing dynamics of an aerial vehicle carrying different payloads. In this approach, the parameters causing the variations in the dynamics are inferred by deploying the policy in the target domain, which in turn helps improve the policy's performance. In Ignasi et al. \cite{ignasi}, the idea of model-agnostic meta-learning \cite{finn2017modelagnostic} was extended to modelling dynamics of a robot. The authors presented an approach to quickly adapt the model of the robot in a new test environment while using a sampling-based controller MPPI to compute the actions. \cite{zeroshotdriving} developed a zero-shot transfer for policy by combining reinforcement learning and a robust tracking controller with a disturbance observer in the target environment. The validated the approach on a vehicle driving task. Similarly, \cite{fan2019bayesian,pmlr-v120-gahlawat20a} presented an approach to combine bayesian learning and adaptive control by learning model error and uncertainty.  

For tasks such as assistive device control for human locomotion, it is potentially unsafe and prohibitive to collect sufficient task-relevant data in the real world which prevents us from using methods such as system identification or transfer learning approaches that need data in the target environment. In addition to this, human dynamics exhibit large variations due to many unobserved parameters, this makes it challenging to define the right parameters for the system model in simulation and also in finding the right range of parameter variation for an approach like DR.

\subsection{Adaptation for Assistive Devices}

Assistive devices such as exoskeletons provide unique challenges for domain adaptation due to the large variations between individuals who pilot the device. Zhang et al \cite{Zhang1280} reported a human-in-the-loop optimization approach for ankle exoskeletons to account for this variability, however, this approach takes a few hours per individual to find the optimal control law. Jackson et al \cite{steve1} presented a unique heuristic-based approach to design a control law that adapts to the person's muscle activity. While these methods work well for steady-state walking, the large number of data required to optimize for in the case of \cite{Zhang1280} and the complex muscle responses involved during push recovery make it an infeasible application.
Several recent works have incorporated a learning-based approach to tackling the problem of adaptation, Peng et al \cite{peng2020data} adopted a reinforcement learning approach to learn assistive walking strategies for Hemiplegic patients, which was tested on real human patients and showed robustness and adaptability. However, it requires online data to update the actor-critic network. This process involves deploying a policy on a patient to collect data, for a task like push recovery it might be challenging to collect relevant data required for updating the policy without compromising the patient's safety. Both \cite{Huang2016AdaptiveIC} and \cite{DMPbasedRL} combined dynamic motion primitives (DMPs) and learning approaches to adapt control strategies for different individuals. Majority of the work with assistive devices have primarily focused on walk assistance and not on push-recovery like this paper addresses.

\begin{comment}
\subsection{Learning based adaptive control}

In \cite{PETS}, showed how to use uncertainty aware dynamic models to increase sample efficiency of RL.

In \cite{ChebotarHZSSL17}, combining model based and model free updates for trajectory centric learning. 

\cite{fan2019bayesian} \textbf{Bayesian learning based adaptive control for safety critical learning}

\cite{pmlr-v120-gahlawat20a} \textbf{L1-GP L1 Adaptive control with bayesian learning}

\cite{ramos2019bayessim} \textbf{BayesSim : Adaptive domain randomization via probabilistic inference for robotics simulation}

\cite{lin2020modelbased} model based adversarial meta RL - they have a lot of gym environments, although their method assumes that the MDP distribution varies in the reward function and the dynamics remains the same.

\cite{fan2019bayesian} - \textbf{Bayesian Learning based adaptive control for safety critical systems}

\cite{gal2015dropout} - \textbf{Dropout as a Bayesian Approximation: Representing Model Uncertainty in Deep Learning}
\end{comment}
\section{Zero-shot Transfer with Error-aware Policy}
% \karen{You use "small commmand" for some captions but not all of them. I think it looks better small. Make sure every caption is in the same format.}
We present a method to achieve the zero-shot transfer of control policies in partially observable dynamical environments. We consider robotic systems and environments with unobservable or unmeasurable model parameters, which make building accurate simulation models difficult. 
%One notable example is an assistive robotic device that needs to interact with a human agent. In this case, an accurate simulation requires obtaining each user's dynamics parameters, such as muscle activation dynamics, tendon layouts, or fatigue models, which are notoriously hard to estimate. Even for robots themselves, there exist many unobservable parameters such as joint slackness or circuit dynamics.

We present a novel policy architecture, an Error-Aware Policy (EAP), that is explicitly aware of errors induced by unobservable dynamics parameters and self-corrects its actions according to the errors. An EAP takes the current state, observable dynamic parameters, and predicted errors as inputs and generates corrected actions. We learn an additional error-prediction function that outputs the expected error. Both the error-aware policy and the error-prediction function, are iteratively learned using model-free reinforcement learning and supervised learning.

\subsection{Problem Formulation}
We formulate the problem as Partially Observable Markov Decision Processes (PoMDPs), $(S, O, A, P, R, \rho_0, \gamma)$, where $S$ is the state space, $O$ is the observation space, $A$ is the action space, $P$ is the transition function, $R$ is the reward function, $\rho_0$ is the initial state distribution and $\gamma$ is a discount factor. 
In our formulation, we make a clear distinction between observable model parameters $\boldsymbol{\mu}$ and unobservable parameters $\boldsymbol{\nu}$ of the agent and environment. Observable quantities are parameters that can be easily measured such as masses or link lengths, whereas unobserved quantities are challenging to estimate, such as circuit dynamics or backlash.
Therefore, both $\boldsymbol{\mu}$ and $\boldsymbol{\nu}$ affect the transition function $P(\mathbf{s'}|\mathbf{a},\mathbf{s},\boldsymbol{\mu},\boldsymbol{\nu})$.
Since we can configure our simulator with both $\boldsymbol{\mu}$ and $\boldsymbol{\nu}$, we can randomly sample $\boldsymbol{\mu}$ and $\boldsymbol{\nu}$ and create a list of  $K$ different environments $\vc{D}=\{(\boldsymbol{\mu}_0, \boldsymbol{\nu}_0), (\boldsymbol{\mu}_1, \boldsymbol{\nu}_1), \cdots, (\boldsymbol{\mu}_K, \boldsymbol{\nu}_K)\}$, but it is hard to obtain $\boldsymbol{\nu}$ at testing time.
In this case, the transition function will be abbreviated as $P(\mathbf{s'}|\mathbf{s},\mathbf{a},\boldsymbol{\mu})$.

Instead of estimating the values of unobserved quantities, we capture the effect of these parameters by defining a metric called a \emph{state-error}. When transferring from one environment to another, the action $\mathbf{a}$ applied at a given state $\mathbf{s}$  will produce different next states due to the differences in both $\boldsymbol{\mu}$ and $\boldsymbol{\nu}$, in other words, a state-error.

% lets say from $\mathbf{P_1}$ to $\mathbf{P_2}$, due to the variations parameterized by $\boldsymbol{\mu}$ and $\boldsymbol{\nu}$, at any given state $\mathbf{s}$ and action $\mathbf{a}$ the next state produced by these two environments will be different, producing a state-error. 
% This is illustrated in Figure \ref{fig:StateError}. 

\begin{figure}
\centering
\setlength{\tabcolsep}{1pt}
\renewcommand{\arraystretch}{0.7}
\vspace{5mm}
\includegraphics[width=0.48\textwidth]{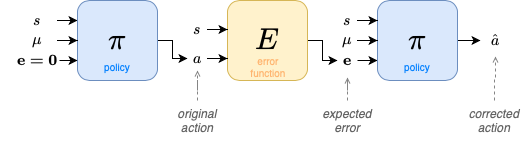}
\vspace{5mm}

\caption{Overview of An Error-aware Policy (EAP). An EAP takes the ``expected'' future state error as an additional input. The expected error is predicted based on the current state $\vc{s}$, observable parameters $\bm{\mu}$, and an uncorrected action $\vc{a}$ that assumes zero error. }
\label{fig:StateError}
\end{figure}

We hypothesize that a policy which is explicitly aware of the state-error would be able to make better decisions by self-correcting its action. We call this an error-aware policy $\pi(\mathbf{a}|\mathbf{s},\boldsymbol{\mu},\mathbf{e})$ (EAP),  which takes in observable parameters $\boldsymbol{\mu}$ as well as the ``expected'' future state error in a new environment $\mathbf{e}$ as input (Figure \ref{fig:StateError}).

%  \sehoon{testing? I think it should be testing.}\karen{I change the wording. See if it is clear now. Sehoon?}  
%To address this \karen{I don't think the following text is addressing the issue raised at the end of the previous paragraph. I would think what you need to say here to address the issue is your meta-learning like training scheme.},
We present a novel training methodology using model-free reinforcement learning that involves learning two functions: an error-aware policy and an error prediction function. First, we learn an error-aware policy that takes the output of error prediction function $E$ as an input and has the ability to generalize to novel environments in a zero-shot manner. Simultaneously, we learn an error-prediction function, which takes as inputs the state $\mathbf{s}$, an uncorrected action $\mathbf{a}$ and observable parameters $\boldsymbol{\mu}$, and outputs the expected state error $\mathbf{e}$ when a policy trained in one environment is deployed to a different one $E: (\mathbf{s}, \mathbf{a}, \boldsymbol{\mu}) \mapsto \mathbb{R}^n$.
We will discuss more details of training in the following sections.
% Similar to meta-learning approaches where the task distribution is split into train-validation sets, we sample a set of \textbf{$n$} different dynamical environments, out of these \textbf{$n$} environments, we randomly choose one environment as a reference dynamical environment $\bar{P}(\mathbf{s'}|\mathbf{s},\mathbf{a},\boldsymbol{\bar{\mu}},\boldsymbol{\bar{\nu}})$ as the training set and the rest \textbf{$n-1$} as a validation set $\tilde{P_{v_i}}(\mathbf{s'}|\mathbf{s},\mathbf{a},\boldsymbol{\tilde{\mu_i}},\boldsymbol{\tilde{\nu_i}})$, where \textbf{$i$} is a single environment in the validation set. The state error $\mathbf{e}$ in a new environment is defined with respect to the reference dynamical model. 

\begin{algorithm}
\caption{Train an Error Aware Policy.}
\label{alg:eap}
\begin{algorithmic}[1]
% \STATE \textbf{Input:} Data-set of $(s,a)$ pairs for n environments : $D_{1..n}$
% \STATE \textbf{Input:} Hold out m (randomly chosen) data as test set : $D_{1..m}$
\STATE \textbf{Input:} Environments $\vc{D}=\{(\boldsymbol{\mu}_0, \boldsymbol{\nu}_0), \cdots, (\boldsymbol{\mu}_K, \boldsymbol{\nu}_K)\}$
% \STATE \textbf{Input:} Reference dynamical environment with $\boldsymbol{\mu}$ \sehoon{how about $\mu_0$?}

\STATE Pre-train $\pi(\mathbf{a}|\mathbf{s},\boldsymbol{\mu}_0,\mathbf{e}=0)$ for $P(\vc{s'}|\vc{s},\vc{a},\boldsymbol{{\mu}}_0)$ reference environment with $\vc{e}=0$
\WHILE {not done}
\STATE Sample an environment with $(\boldsymbol{\mu}, \boldsymbol{\nu})$ from $\vc{D}$
\FOR{ each policy update iteration}
\STATE Initialize buffer $\vc{B}=\{\}$ 
\STATE Update an error function $E$ using Algorithm~\ref{alg:errorfunc}
\STATE $\vc{B}$ = Generate rollouts using Algorithm~\ref{alg:rollout} 
\STATE Update policy $\pi$ using $\vc{B}$ with PPO.
%\ENDFOR

\ENDFOR \label{line:trainingSetEnd}
\ENDWHILE
\RETURN{ $\pi(\mathbf{a}, |\mathbf{s}, \boldsymbol{\mu}, \mathbf{e})$}
\end{algorithmic}
\end{algorithm}
\subsection{Training an Error-aware Policy}
\noindent\textbf{Training Procedure.} The training process of an error-aware policy is summarized in Algorithm~\ref{alg:eap}.
Assume that we have an oracle error function $E(\mathbf{s}, \mathbf{a}, \boldsymbol{\mu})$ that outputs the expected state error in a novel environment, which will be explained in the following section. First, the policy is pre-trained to achieve the desired behavior only in the reference environment $(\boldsymbol{\mu}_0, \boldsymbol{\nu}_0)$ assuming there is no state error, $\pi(\mathbf{a}|\mathbf{s},\boldsymbol{\mu}_0,\mathbf{e}=0)$. Once the policy is trained in the reference environment, we sample dynamics parameters $\boldsymbol{\mu}_i$ and $\boldsymbol{\nu}_i$ ($i>0$) uniformly from the data set $\vc{D}$  and evaluate the EAP in this new environment. The policy parameters are updated using a model-free reinforcement learning algorithm, Proximal Policy Optimization~\cite{schulman2017proximal}.
Sampling new testing environments and updating policy parameters are repeated until the convergence.

\begin{algorithm}
\caption{Generate Rollouts}
\label{alg:rollout}
\begin{algorithmic}[1]
\STATE \textbf{Input:} Observable dynamics parameters $\bm{\mu}$, Transition function $P$, Current policy $\pi$ and error function $E$, Replay buffer $\vc{B}$ %\karen{What $\bm{\nu}$ is used in $P$?} \sehoon{In simulation, we can use the corresponding parameter. In real-world experiment, we don't need $\bm\nu$ for running the experiment.}\karen{Should we pass $\nu$ in just like $\mu$?}\sehoon{Maybe. But I want to avoid the impression that we need to know the exact values of $\bm\nu$. It might be too subtle though.}
\STATE Sample state $\vc{s}$ from initial state distribution $\rho_0$ 
\WHILE{not done}
\STATE $\vc{a} \sim \pi(\vc{a}|\vc{s},\bm{{\mu}},\vc{e}=0)$ \tcp{original action}
\STATE $\vc{e} = E(\vc{s}, \vc{a}, \bm{{\mu}})$ \tcp{predicted error}
\STATE $\hat{\vc{a}} \sim \pi(\vc{a}|\vc{s},\bm{{\mu}},\vc{e})$ \tcp{error-aware action}
\STATE $\vc{s}'\sim  P(\vc{s}'|\vc{s}, \hat{\vc{a}}, \bm{\mu})$ 
\STATE $r = R(\vc{s},\hat{\vc{a}})$ 
\STATE $B = B \cup \{(\vc{s},\hat{\vc{a}},r,\vc{s}',\bm{\mu})\}$
% \STATE Predict error $\vec{e}=f(\phi)$ with inputs $s,a,\mu_{o}$ 
% \STATE Predict action to be taken with $\pi(a|s,\mu_{o},e)$
% \STATE Take one simulation step with  $s'=P_{all}(s'|s,a,\mu_{o})$
% \STATE Compute reward with $r_{task}(s,a)$
% \STATE Store tuple $(s,a,r,s')$ in buffer $B_{pol}$
\STATE $\vc{s}=\vc{s}'$
 
\ENDWHILE
\RETURN $B$
\end{algorithmic}

\end{algorithm}

\noindent\textbf{Rollout Generation.}
A roll-out generation procedure is described in Algorithm~\ref{alg:rollout}.
Given a state $\mathbf{s}$ in this environment $(\boldsymbol{\mu}, \boldsymbol{\nu})$, we query an action from policy $\pi$ as if the policy is being deployed in the reference environment with $\mathbf{e} = 0$. This action $\vc{a}$  is fed into the error function $E$ which predicts the expected state error in this environment, then the state error is  passed into the error-aware policy to query a corrected action $\hat{\vc{a}}$ which will be applied to the actual system. The task reward $R(\vc{s},\vc{a})$ guides the policy optimization to find the best ``corrected'' action that maximizes the reward.

\begin{algorithm}
\caption{Train an Error Prediction Function.}
\label{alg:errorfunc}
\begin{algorithmic}[1]
% \STATE \textbf{Input:} $\tilde{\mu} \sim D_{val}$
\STATE \textbf{Input:} Reference environment with $\bm{\mu_0}$
\STATE \textbf{Input:} Target environment with $\bm{\mu}$
\STATE \textbf{Input:} Replay Buffer $\vc{B}$
\STATE \textbf{Input:} Dataset $\vc{Z}$
\STATE \textbf{Input:} Error Horizon $T$
% \karen{You call this Augmented Replay Buffer like it is an augmentation of $B$. Is it true?}
%\STATE Initialize a training buffer $B = \{\}$ 
% \FOR{$i = 1:K$}
% \WHILE{not converged\sehoon{Do we need this? What should be converged?}} 
 %\STATE Initialize state $\vc{\tilde{s}}^0$ and $\vc{\bar{s}}^0, $ 
 %\STATE Initialize state $\vc{s}$.  \sehoon{I guess we only need one initial state?}
 %\STATE $\vc{s}$ = a state after running a policy $\pi(\vc{a}_{t}|\vc{s}_t,\bm{\mu}_{o}, \vc{e}=0)$ on $P$ for T steps.

\WHILE{not done}
\STATE Sample the initial state $\vc{s}_0^0$ from $\vc{B}$
% \karen{What is $\rho_0$? You define initial distribution as $p_0$ in the text.}
\STATE $\vc{s}^0 = \vc{s}_0^0$
% \STATE Set $\vc{s}$ in  $P$ (denoted $\vc{s}$) and $\bar{P}$ (denoted $\bar{\vc{s}}$) \karen{$P$ is the transition function as you defined earlier. I don't understand how it can take a state $\vc{s}_0$ as input. Do you just mean $\vc{s}^0 = \vc{s}$ and $\bar{\vc{s}}^0 = \vc{s}$? And what is $\vc{s}^0_{v_i}$?}

\FOR{$t = 0:T-1$}
\STATE \text{}\tcp{Simulation in Reference Env}
\STATE $\vc{a}^{t}_0 \sim \pi(\vc{a}|\vc{s}_0^t,\bm{\mu}_0, \vc{e}=0)$ 
% \karen{$\vc{a}_t$ is never used in the algorithm as the pseudo code presents. Is $\vc{a}_t$ different from $\vc{a}^t$?} 
% \tcp{action in the reference env} 
\STATE $\vc{s}^{t+1}_0 \sim P(\vc{s}^t_0, \vc{a}^{t}_0 ,\bm{\mu}_0)$
% \STATE $\vc{s}'  \leftarrow$ Step($\vc{s},\vc{a}$) in environment $P$ \karen{P is not a distribution of state. It is a transition function.}
% \STATE $\vc{\bar{a}} \sim \pi(\vc{\bar{a}}|\vc{\bar{s}},\bm{{\mu_0}}, \vc{e}=0)$ \karen{Same issues as Line 7.}%\tcp{action in the validation env} 
% \STATE $\vc{\bar{s}'} \leftarrow$ Step($\vc{\bar{s}},\vc{\bar{a}}$) in environment $\bar{P}$ \karen{Same issues as Line 8.}

% \STATE $\vc{s}$ = $\vc{s}'$
% \STATE $\bar{\vc{s}}$ = $\bar{\vc{s}}'$
% \STATE $\vc{a_0}$ $\leftarrow$ Store first action ,$\vc{a}$ at $t=0$
% \STATE $\vc{s_0}$ $\leftarrow$ Store first state ,$\vc{s}$ at $t=0$

\STATE \text{}\tcp{Simulation in Validation Env}
\STATE $\vc{a}^{t} \sim \pi(\vc{a}|\vc{s}^t,\bm{\mu}, \vc{e}=0)$ 
\STATE $\vc{s}^{t+1} \sim P(\vc{s}^t, \vc{a}^{t}, \bm{\mu})$

\ENDFOR
\STATE $\vc{Z} = \vc{Z} \cup \{(\vc{s^0},\vc{a^0},\vc{s}^T, \vc{s}_0^T,\bm{\mu})\}$
% \STATE $\vc{s}'_{r,t+1} \sim P_{r}(\vc{s}'|\vc{s},\vc{a},\bm{\mu}_{o})$  
%\ENDFOR
% \STATE $\vc{e}=\vc{{s}}^{T}-\vc{\tilde{s}}^{T}$
% \STATE $B = B \cup \{(\vc{s}^0,\vc{a}^0,\bm{\tilde{\mu}},\vc{{s}}^{T},\vc{\tilde{s}}^{T}$)\}
% \STATE Store tuple $(s,a,\mu_{o},\vec{e})$ in buffer $\mathcal{B}$
\ENDWHILE
% \ENDFOR
\STATE minimize the $L(\phi)$ in Eq.~\ref{eq:error_prediction_loss} using $\vc{Z}$.
% \STATE Train neural network $\phi$, with inputs $(s,a,\mu_{o})$ and output $\vec{e}$ using supervised learning with loss $L(\phi)=||f(\phi)-e||^{2}$ 

\RETURN $\phi$
\end{algorithmic}

\end{algorithm}

\subsection{Training an Error Function} \label{sec:error_func}
In reality, we do not have an oracle error function that can predict the next state due to the lack of unobservable parameters $\boldsymbol{\nu}$. To this end, we will learn this function simultaneously with EAP, by splitting the dataset $\vc{D}$ into the training and validation sets. Similar to training methodology followed in meta-learning algorithms, we repeatedly apply the trained policy into sampled environments from the validation set. Because our nominal behavior is pre-trained in the reference environment $(\boldsymbol{\mu}_0, \boldsymbol{\nu}_0)$, we compute the errors by measuring the differences in the reference environment $(\boldsymbol{\mu}_0, \boldsymbol{\nu}_0)$ and the validation environment $(\boldsymbol{\mu}, \boldsymbol{\nu})$: $\bm{e} = (\vc{\bar{s}}' - \vc{s}') \in \mathbb{R}^n$, generated by two dynamic models ${P}(\vc{s}'|\vc{s}, \vc{a}, \bm{{\mu}}_0)$ and $\bar{P}(\vc{s}'|\vc{s},\vc{a}, \bm{\mu})$. 

\noindent \textbf{Horizon of Error Prediction.} In practice, we found that the error accumulated during one step is often not sufficient to provide useful information to the EAP. To overcome this challenge, we take the state in the collected trajectory and further simulate it for $T$ steps in both the reference environment ${P}(\vc{s}'|\vc{s}, \vc{a}, \bm{{\mu}}_0)$ and the validation environment $\bar{P}(\vc{s}'|\vc{s},\vc{a}, \bm{\mu})$. 
% we find that the state-error after a horizon length $T$ provides  more information about how different the environment is compared just one step state error ($T=1$). 
We provide analysis on the effect of horizon length from $T=1$ to $T=8$ in the Section \ref{sec:experiment}. 
% The hyperparameter $T$ in Algorithm \ref{alg:errorfunc} represents the horizon length. 
%Note that we matched the values of the observable parameters $\bm{\tilde{\mu}}$ that can be easily identified. \sehoon{Are we doing this?}

\noindent \textbf{Loss Function.} Since the differences between the two dynamical environments reflects the reality gap caused by unobservable parameters, the error prediction function $E$ enables us to learn the effect of the unobserved parameters captured through the state error. We train our error prediction function $E$ to learn this ``emulated'' sim-to-real gap by minimizing the following loss:
\begin{equation} \label{eq:error_prediction_loss}
    L(\bm{\phi}) = \sum_{(\vc{s^0},\vc{a^0},\vc{s}^T, \vc{s}_0^T,\bm{\mu})\in \vc{Z}}  ||E(\vc{s}^0, \vc{a}^0,\bm{\mu})  - (\vc{s}_0^T - \vc{s}^T)||^2,
\end{equation}
where $\vc{Z}$ is the collected dataset and $\bm{\phi}$ is the parameters or the neural net representing $E$.
% where $\vc{s'}^{T}$ and $\vc{s'}_{v_i}^{T}$ are the future state after $T$ steps, ${P}$ and $\tilde{P_{v_i}}$, respectively. 
Algorithm~\ref{alg:errorfunc} summarizes the training procedure.
% As the training progresses, the error function improves its predictive capability.

\begin{figure}
\centering
\setlength{\tabcolsep}{1pt}
\renewcommand{\arraystretch}{0.7}
  \includegraphics[width=0.48\textwidth]{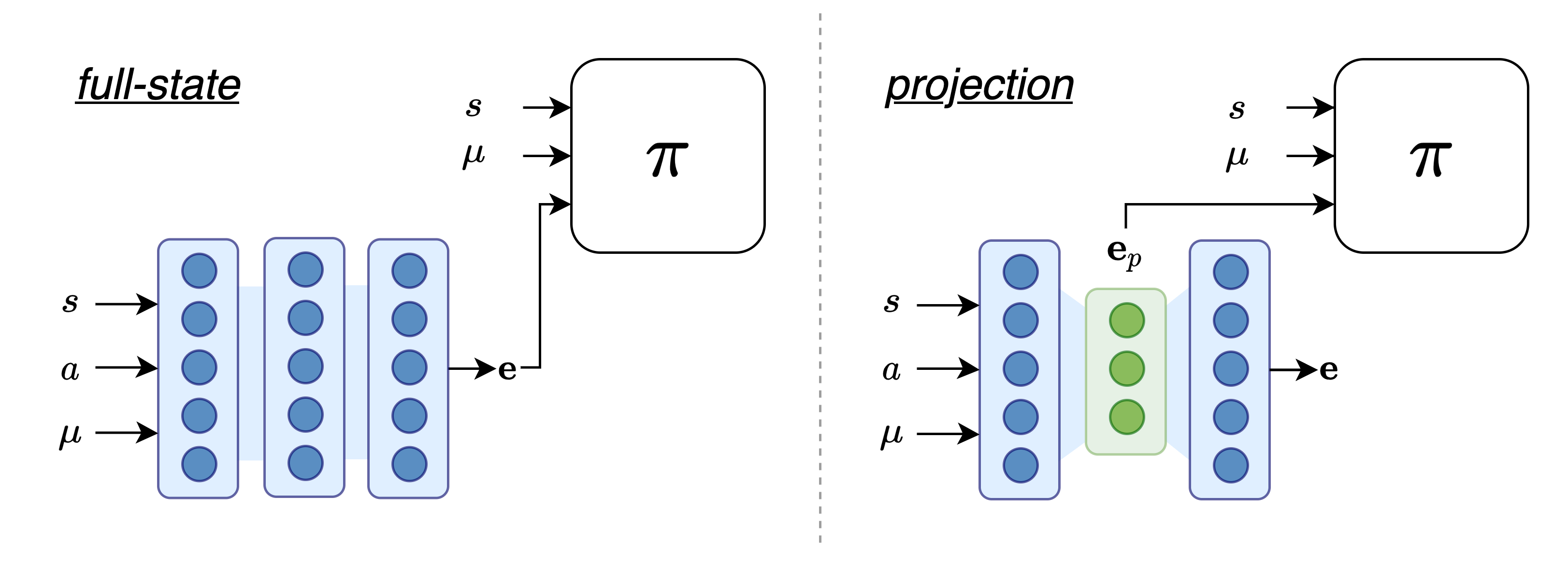}
\caption{\small{\textbf{Left :} A full state error representation input into the policy vs \textbf{Right :} Projected error representation as an input to the policy }}
\label{fig:ErrorRep}
\end{figure}

\noindent \textbf{Reduced Representations.} We experiment with two different representations of the error input to the policy. First, we input the full state error $\vc{e}=\vc{s}_0^T - \vc{s}^T$(with the same dimension as the state) approximated by a MLP neural network, into the policy. Second, we use a network architecture with an information bottle neck, as illustrated in Figure \ref{fig:ErrorRep}, and input the latent representation $\vc{e}_p$ 
% \karen{Why introduce another symbol that is not used later?}\visak{I use this later in the experiment section to talk about the projected dimension} 
into the policy. The same loss function $L$ is used to train both the functions.

\section{EVALUATION}
\label{sec:experiment}

We design experiments to validate the performance of error-aware policies. We aim to answer the following research questions.
\begin{enumerate}
    \item Does an EAP show better zero-shot transfer on unseen environments compared to the baseline algorithms?
    \item How does the choice of hyperparameters affect the performance of an EAP? 
\end{enumerate}

% We evaluate the performance of error-aware policies on five different tasks, first we learn a push-recovery control policy using an assistive device for simulated human agents. In addition, we validate our method on a few simulated agents performing tasks defined by the OpenAI Gym environments such as cartpole, hopper, walker 2D and a quadruped task using Unitree's A1 robot. Our experiments aim to answer the following questions:
% \begin{enumerate}
%     \item For each task, how well does EAP perform in both training and testing environments compared to the baselines? 
%     \item How does the horizon of predicted future state deviation affect the performance of EAP?
%     \item How does a learned representation of the future state deviation compare to the direct difference between two future states?    
%     \item How sensitive is EAP to the choice of observable $\bm{\mu}$?
%     \item How sensitive is EAP to the choice of reference dynamics $\bm{\mu_0}$
% \end{enumerate}

\subsection{Baseline Algorithms}
We compare our method with two baselines commonly used for sim-to-real policy transfer, Domain Randomization (DR)\cite{openai2018learning,Pengdr} and Universal Policy (UP) \cite{yu2017preparing}.
DR aims to learn a more robust policy for zero-shot transfer, by training with randomly sampled dynamics parameters (in our case, both $\bm{\mu}$ and $\bm{\nu}$). UP extends DR by taking dynamics parameters as additional input. UP often transfer to target environments better than DR, but it explicitly requires to know dynamics parameters, where $\bm{\nu}$ is assumed to be unobservable in our scenario.
We did not compare EAPs against meta-learning algorithms \cite{belkhale2020modelbased,finn2017modelagnostic,ignasi}, which require additional samples from the validation environment.

% \begin{enumerate}
%     \item Domain randomization (DR): We train a policy in with domain randomization on the observable model parameters $\bm{\mu}$.
%     \item Universal policy (UP): Like DR, we train a policy with domain randomization on $\bm{\mu}$. In addition, the policy takes in $\bm{\mu}$ as part of the input.
% \end{enumerate}

\subsection{Tasks}

\begin{table*}
\centering
\begin{tabular}{|l|l|l|l|l|}
\hline
Task & Observable Params. $\bm\mu$ & Unobservable Params. $\bm\nu$  & Net. Arch. & Err. Dim. $|\vc{e}_p|$         \\ \hline

Assitive Walking  & mass, height, leg length, and foot length & joint damping, max torques, PD gains and delay &  (64, 32) & 6         \\ \hline
Aliengo & PD gains, link masses & sensor delay, joint damping, ground friction &  (64, 32) & 6         \\ \hline
Cartpole & pole length, pole mass, cart mass  & joint damping, joint friction &   (32, 16) & 2     \\ \hline
Hopper & thight mass, foot mass, shin length & joint damping, ground friction &   (32, 16) & 4       \\ \hline
Walker 2D & link masses, shin length & sensing delay, joint damping, ground friction & (64, 32) &   5        \\ \hline

\end{tabular}
\caption{Tasks and Network Architectures 
% \sehoon{fill the numbers} \karen{We can just say the net architecture somewhere in the text since it is the same for all tasks. Same thing for error horizon (I believe they are all 5). If we remove two columns, can you try to fit the table in a single column?} \sehoon{Explain parameters using verbal words. Then it may become two-column width anyway.}
}
\label{tab:params}
\end{table*}

We evaluate the performance of error-aware policies on five different tasks. The first task is about push-recovery of an assistive walking device for simulated humans, inspired by the work of Kumar et al~\cite{kumar2019learning}.
The second task is locomotion of a quadrupedal robot, Aliengo Explorer\cite{UnitreeAliengo}.
The rest three tasks are CartPole, Hopper, and Walker2D, which are from the OpenAI benchmark suite \cite{GymAI}.

\subsubsection{Assistive walking device for push recovery}
\begin{figure}
\centering
\setlength{\tabcolsep}{1pt}
\renewcommand{\arraystretch}{0.7}
\includegraphics[width=0.4\textwidth,height=5cm]{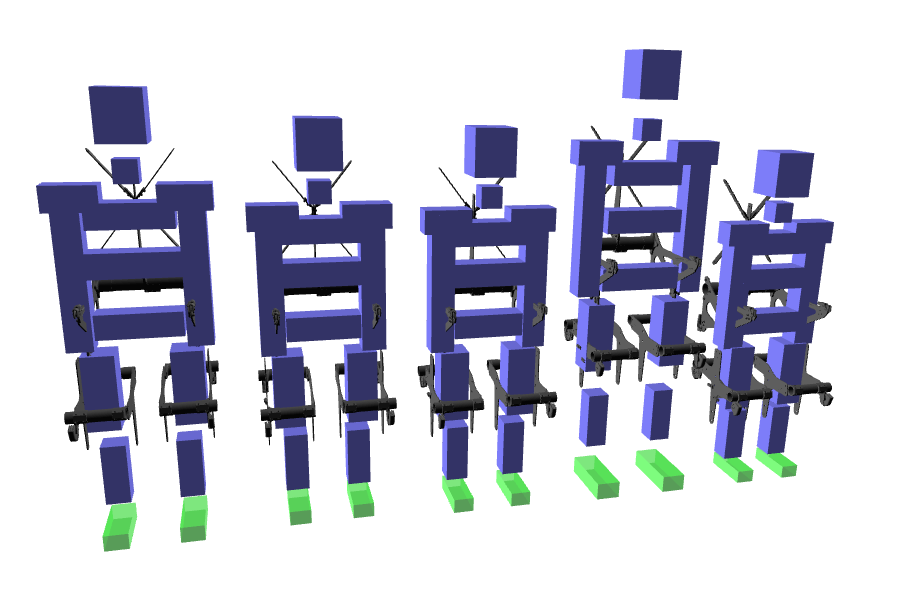}
\caption{Five different test subjects for the assistive walking experiment with varying height, mass, leg length and foot length from the biomechanical gait dataset \cite{dataset2018}.}
\label{fig:TS}
\end{figure}

In this task, the goal is to learn a policy for an assistive wearable device (i.e. exoskeleton) to help a human recover balance after an external push is applied.
(inset figure).
% \begin{wrapfigure}{r}{0.2\textwidth}
%     \centering
%     \includegraphics[width=0.2\textwidth]{images2/EXO.png}
% \end{wrapfigure}
We use a hip exoskeleton that applies torques in 2-degrees of freedom at each hip joint. Our algorithm begins by training $15$ human agents using public biomechanical gait data \cite{dataset2018} to walk in a steady-state gait cycle, similar to the approach presented in \cite{peng2018deepmimic}. The $15$ agents vary in mass, height, leg length, and foot length according to the biomechanical data used to train their corresponding policies, which formulate the four-dimensional observable parameters $\bm{\mu}$ (Figure~\ref{fig:TS}). We also vary each human agent's joint damping, maximum joint torques, PD gains, and sensory delay as the four dimensional unobservable parameters $\bm{\nu}$. We split the $15$ human agents into $10$ for the training set and $5$ as the testing set.

\noindent \textbf{Human Behavior Modeling.}
First, we capture the human behavior by training a human-only walking policy $\pi_h$ that mimics the reference motion which frames are denoted as $\bar{\vc{q}}$.
Each human model has $23$ actuated joints along with a floating base. The state space has $53$ dimensions, $\vc{s}_{h} = [\vc{q},\vc{\dot{q}},\vc{v}_{com}, \boldsymbol{\omega}_{com},\psi]$, which represent joint positions, joint velocities, linear and angular velocities of the center of mass, and a phase variable $\psi$ that indicates the target frame in the reference biomechanical gait cycle. 
% We model the intrinsic sensing delay of a human musculoskeletal system by adding a latency of $40$ milliseconds to the state vector before it is fed into the policy. \sehoon{too much details..}
The action $\vc{a}$ is defined as the offset to the reference biomechanical joint trajectory $\bar{\vc{q}}(\psi)$, which results in the target angles: $\vc{q}^{target} = \bar{\vc{q}} + \vc{a}$.
% \karen{What is the action? $\vc{a}$ or $\vc{q}^{target}$? The current writing seems to define action as $\vc{q}^{target}$. Then what is $\vc{a}$?}\sehoon{Updated.} 
The reward function encourages mimicking the reference motions from public biomechanical data: %\karen{Here you use subscript for time, but in other places, you use superscript for time. Make it consistent. I recommend to use superscript for time everywhere.}\sehoon{I agree, but maybe very time-consuming. It might be practical to remove subscripts from the above...} \karen{Agree. I think it is the only place where t is subscript.}:
\begin{multline}
    \label{eqn:reward}
    R_{human}(\vc{s}_{h},\vc{a}_{h}) =
    w_{q}(\vc{q} - \bar{\vc{q}}) + w_{v}(\vc{\dot{q}} - \vc{\bar{\dot{q}}})\\ 
    + w_{c}(\vc{c} - \bar{\vc{c}}) + w_{p}(\vc{p} - \bar{\vc{p}}) - w_{\tau}||\boldsymbol{\tau}||^{2},
\end{multline}
where the terms include the reference joint positions $\vc{\bar{q}}$, joint velocities $\vc{\bar{\dot{q}}}$, end-effector locations $\vc{\bar{p}}$, contact flags $\vc{c}$, and the joint torques $\bm{\tau}$. During training,  we exert random forces to the agent during policy training. Each random force has a magnitude uniformly sampled from $[0,800]\ N$ and a direction uniformly sampled from [-$\pi/2$,$\pi/2$], applied for $50$ milliseconds on the agent's pelvis in parallel to the ground. The maximum force magnitude induces a velocity change of roughly $0.6 - 0.8$ m/sec. This magnitude of change in velocity is comparable to experiments found in biomechanics literature such as \cite{Wang2014},\cite{Agarwal} and \cite{hof2010balance}. We also randomize the time when the force is applied within a gait cycle. The forces are applied once at a randomly chosen time in each trajectory rollout.
% For more details, please refer to the work of Visak et al~\cite{kumar2019learning}.

\noindent \textbf{MDP Formulation.}
Once the human agents are trained, we begin learning the push-recovery EAP for the assistive device.
The objective is to stabilize the human gait from external perturbations.
The $17$ dimensional state of robot is defined as $\vc{s}_{e}=[{\bm{\omega}},\bm{\alpha},\vc{\Ddot{x}},\vc{q}_{hip},\dot{\vc{q}}_{hip}]$, which comprises angular velocity, orientation, linear acceleration, hip joint positions, and hip joint velocity.
The four dimensional action $\vc{a}_e$ consists of torques at two hip joints.
The reward function maximizes the quality of the gait while minimizing the impact of an external push.
\begin{multline}
    \label{eq:exoReward}
    R_{exo}(\vc{s}_h, \vc{s}_e, \vc{a}_e) = R_{human}(\vc{s}_h) \\ - w_{1}\|\vc{v}_{com}\| - w_{2}\|\boldsymbol{\omega}_{com}\| - w_3 \|\vc{a}_e\|,
\end{multline}
where $R_{human}$ is defined in equation \ref{eqn:reward} , and $\vc{v}_{com}$ and $\boldsymbol{\omega}_{com}$ are the global linear and angular velocities of the pelvis. The last term penalizes the torque usage. We use the same weight $w_{1} = 2.0$, $w_{2} = 1.2$ and $w_{3} = 0.001$ for all our experiments. 
% \karen{People will ask why Equation 3 takes as input $\vc{s}_h$ not $\vc{s}_e$. You need to explain that.}

% Once an external push is applied to the human agent, the exoskeleton policy is activated and the torques continue to be applied for the rest of the trajectory.   The exoskeleton policy $\pi_e(\vc{a}_e|\vc{s}_e,\bm{\mu}_o,\vc{e},)$ also takes the observable parameters of the human agent $\bm{\mu}$ as input. 

% \noindent \textbf{Policy Architecture.}
% The EAP is represented as a MLP neural network with two hidden layers with 128 neurons. The error function $E$ takes as input the state of dimension $\vc{s} \in R^{17}$ and projects it down to $R^{5}$, as described in Section~\ref{sec:error_func}. 

% The set of observable parameters $ \bm{\mu} \in R^{4}$ comprises of the subjects  mass, leg-length, height and foot length. The unobservable parameters include $ \bm{\nu} \in R^{4}$ joint damping, joint torque limits, sensing delay and the PD controller gains for the human agent. 

% : assistive device walking, quadrupedal walking,  first we learn a push-recovery control policy using an assistive device for simulated human agents. In addition, we validate our method on a few simulated agents performing tasks defined by the OpenAI Gym environments such as cartpole, hopper, walker 2D and a quadruped task using Unitree's A1 robot.

\subsubsection{Quadrupedal Locomotion}
In our second task, we learn a control policy that generates a walking motion for a quadrupedal robot, Aliengo Explorer \cite{UnitreeAliengo}. For this task, the $17$ observable parameters ($\bm{\mu}$) are PD gains of the joints, link and root masses and the $10$ unobservable parameters $\bm{\nu}$ include sensing delay, joint damping of thigh and knee joints and ground friction. The 39-dimensional state space consists of torso position and orientation and corresponding velocities, joint position and velocities, foot contact variable that indicates when each foot should be in contact with the ground, while the $12$-dimensional action space consists of joint velocity targets which is fed into a PD controller that outputs torques to each joint.

The reward function is designed to track the target motion that walks at $0.8$~m/s:
\begin{multline}
    r(\vc{s},\vc{a}) = w_1 e^{-k_{1}*(\vc{q} - \vc{\bar{q}})} + w_2 e^{-k_{2}*(\vc{\dot{q}} - \vc{\bar{\dot{q}}})} \\ +  w_3 \min(\dot{x},0.8) + \sum_{i=1}^4 ||c_i - \bar{c}_i||^2.
\label{eqn:quad}
\end{multline}
In this equation, the first term encourages to track the desired joint positions, the second term is to track the desired joint velocities, the third term is for matching the forward velocity $\dot{x}$ to a target velocity of $0.8$ m/s. and the four term tracks the predefined contact flags.
% \sehoon{Then can you update the formulation of the third term? maybe use the exponential function, like other terms.}
We use the same weight $k_1=35$,$w_1=0.75$, $k_2=2$,$w_2=0.20$, and $w_3=1.0$ for all experiments.

\subsubsection{OpenAI Environments}
We test our method on three OpenAI environments: CartPole, Hopper and Walker2D.
While using the same state spaces, action spaces, and the reward functions described in the benchmark \cite{GymAI}, we additionally define observable and unobservable dynamics parameters as follows: 

% For each task, we need to first manually decide observable and unobservable variations $\bm{\mu} = (\bm{\mu}, \bm{\nu})$ interacting with the target environment. The model parameters $\bm{\mu}$ are listed in Table \ref{tab:mu}. Then we follow EAP procedure to train a policy that can be tested on unseen environments. The results in Figure \ref{fig:TestPerf} shows that our method outperforms the baselines in all three tasks.

\begin{enumerate}
% \item \noindent\textbf{Cartpole.} $\bm{\mu} \in R^3$ includes length of pole $l_{pole}$, mass of cart and pole $m_{cart},\ m_{pole}$. Unobservable parameters $\bm{\nu} \in R^3$ include joint damping $\bm{\beta}$ and joint friction $\bm{\mu_j}$. \karen{Reuse symbol $\bm{\mu}$ for a parameter of $\bm{\nu}$ is a bad idea. Also, why boldface $\beta$ and $\mu_j$? Are they vectors? Also also, why use a subscript $j$?} 
\item \noindent\textbf{Cartpole.} Observable parameters $\bm{\mu} \in R^3$ includes the length of the pole, the mass of the pole, and the mass of cart. Unobservable parameters $\bm{\nu} \in R^3$ include the damping at the rotational joint, the friction at the rotational joint, and the friction at the translational joint. 
% $\bm{\mu_j}$. \karen{Reuse symbol $\bm{\mu}$ for a parameter of $\bm{\nu}$ is a bad idea. Also, why boldface $\beta$ and $\mu_j$? Are they vectors? Also also, why use a subscript $j$?} 

% \item \noindent\textbf{Hopper.} $\bm{\mu} \in R^3$ includes mass of thigh and foot $m_{thigh},\ m_{foot}$ and length of shin bodynode $l_{shin}$. Unobservable parameters $\bm{\nu} \in R^3$ include joint damping of shin and foot joints $\bm{\beta_{shin}},\ \bm{\beta_{foot}}$ and ground friction $\bm{\mu_{gr}}$ \karen{Same comments as above.}.

\item \noindent\textbf{Hopper.} Observable parameters $\bm{\mu} \in R^3$ include the mass of the thigh and foot and the length of the shin bodynode. Unobservable parameters $\bm{\nu} \in R^3$ include joint damping of shin and foot joints and ground friction.

\item \noindent\textbf{Walker 2D.} Observable parameters $\bm{\mu} \in R^6$ include the masses of thigh and foot for both legs, the mass of pelvis, and the length of shin. Unobservable parameters $\bm{\nu} \in R^4$ include joint damping of foot joints, the delay in observation, and ground friction.
\end{enumerate}

\subsection{Zero-shot Transfer with EAPs}

\begin{figure}
\centering
\begin{subfigure}[b]{0.475\columnwidth}
    \centering
    \includegraphics[width=\columnwidth]{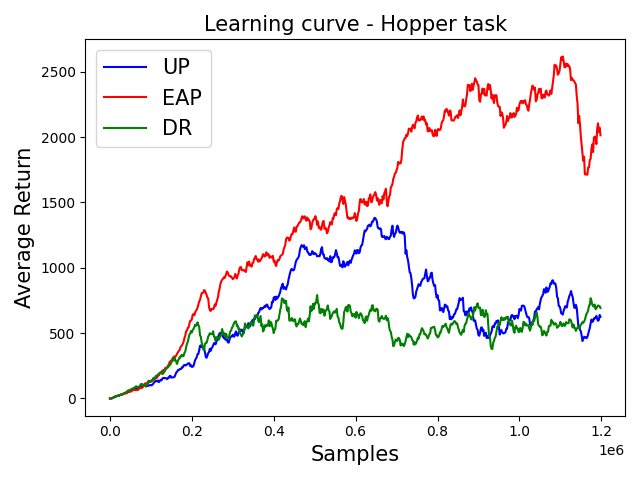}
    \caption[Network2]%
    {{\small Hopper}}    
    \label{fig:mean and std of net14}
\end{subfigure}
\hfill
\begin{subfigure}[b]{0.475\columnwidth}  
    \centering 
    \includegraphics[width=\columnwidth]{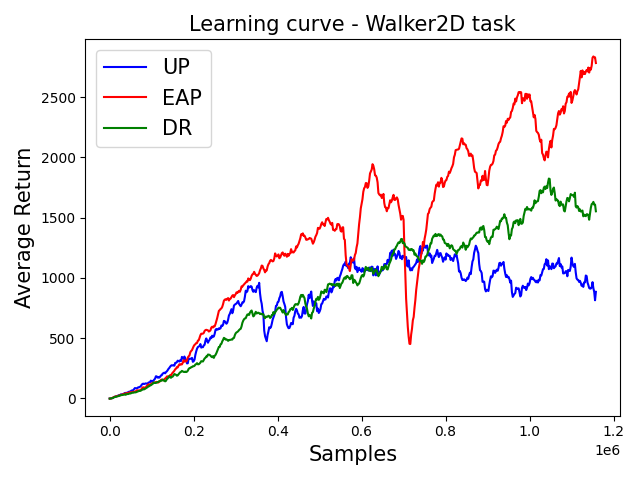}
    \caption[]%
    {{\small Walker 2D}}    
    \label{fig:mean and std of net24}
\end{subfigure}
\vskip\baselineskip
\begin{subfigure}[b]{0.475\columnwidth}   
    \centering 
    \includegraphics[width=\columnwidth]{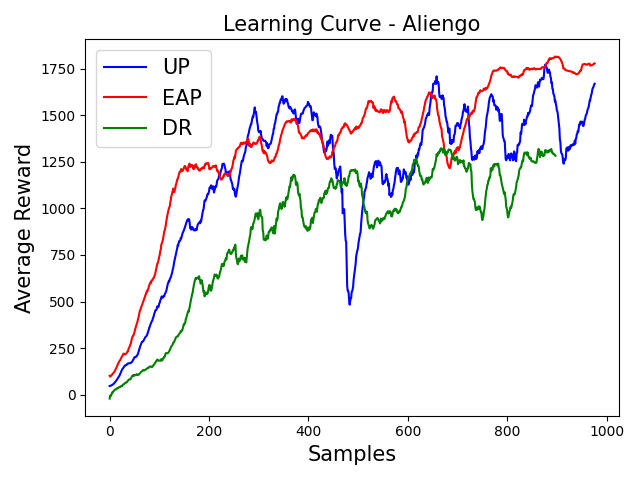}
    \caption[]%
    {{\small Quadrupedal Locomotion}}    
    \label{fig:mean and std of net34}
\end{subfigure}
\hfill
\begin{subfigure}[b]{0.475\columnwidth}   
    \centering 
    \includegraphics[width=\columnwidth]{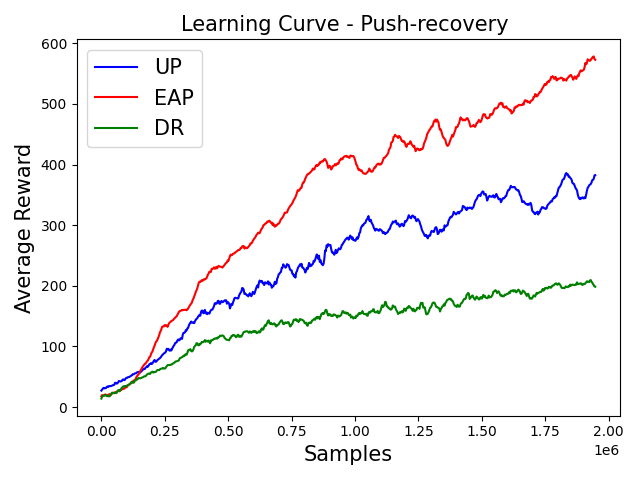}
    \caption[]%
    {{\small {Assistive Walking}}}    
    \label{fig:mean and std of net44}
\end{subfigure}
\caption{\small{ Learning curves for four tasks. The number of samples for EAP include the ones generated for training an error function.}} 
\label{fig:learning_curves}
\end{figure}

In this section, we compare the zero-shot transfer of error-aware policies against two other baseline algorithms, Domain Randomization (DR) and Universal Policies (UP).

\noindent \textbf{Learning Curves.} 
First, we compare the learning curves of the EAP, DR, and UP approaches on four selected tasks in Figure~\ref{fig:learning_curves}.
We set the same ranges of the observable and unobservable parameters for all three algorithms.
In our experience, EAPs learn faster than DR and UP for three tasks, the Hopper, Walker2D, and assistive walking tasks, while showing comparable performance for the quadrupedal locomotion task.
Note that, to make the comparison fair to baselines, we also include the samples for training error functions (Algorithm~\ref{alg:errorfunc}) when we evaluate the performance of EAPs. We do not include the experiment on the CartPole environment for brevity but the EAP outperforms the baselines as well %\karen{People might point out that the learning didn't seem to converge yet for many curves.}. \sehoon{yeah, but nothing we can do about it at this point..}

\noindent \textbf{Zero-shot Transfer.} 
\begin{figure}
\centering
\setlength{\tabcolsep}{1pt}
\renewcommand{\arraystretch}{0.7}
\includegraphics[width=0.4\textwidth,height=5cm]{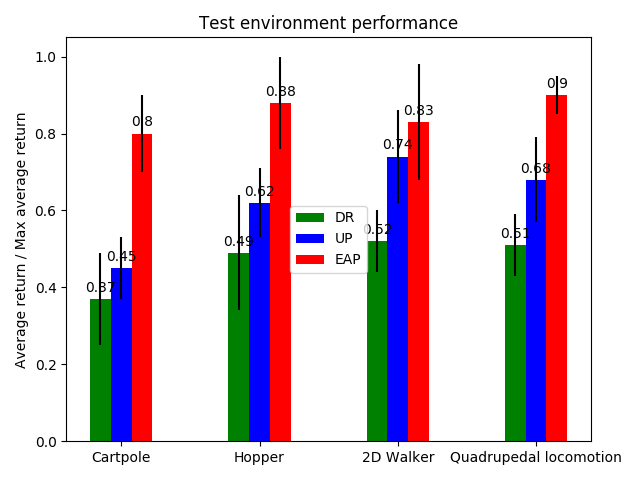}
\caption{\small{Comparison of EAP and baselines DR and UP. The error bars represent the variation in the average return of the policy in the target environment when trained with 4 different seeds.}}
\label{fig:TestPerf}
\end{figure}
Then we evaluate the learned policies on unseen validation environments, where their dynamics parameters $\bm\mu$ and $\bm\nu$ are sampled from the outside of the training range.
We conduct the experiments for the CartPole, Hopper, Walker2D and quadrupedal locomotion tasks and compare the \emph{normalized} average returns (the average return divided by the maximum return).
The results are plotted in Figure~\ref{fig:TestPerf}, which indicate that EAP outperforms DR by $60$\% to $116$\% and UP by $12$\% to $77$\%.
Note that UP may perform well for the real-world transfer due to the lack of the unobservable parameters.
We also observe that UP is consistently better than DR by being aware of the dynamics parameters, $\bm\mu$ and $\bm\nu$, which meets our expectation.

\begin{figure}
\centering
\setlength{\tabcolsep}{1pt}
\renewcommand{\arraystretch}{0.7}
\includegraphics[width=0.4\textwidth,height=5cm]{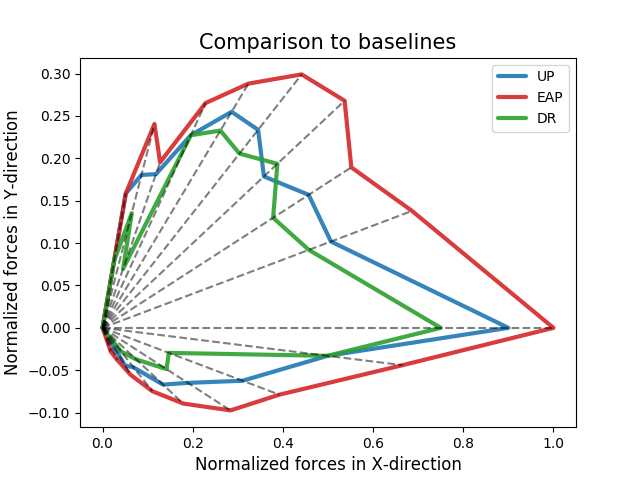}
\caption{Average stability region in five test subjects. The results indicate the better zero-shot transfer of EAP over DR and UP. }
\label{fig:SR_test}
\end{figure}

For evaluating the zero-shot transfer for the assistive walking task, we define an additional metric ``stability region'', which depicts the ranges of maximum perturbations in all directions that can be handled by the human with the EAP-controlled exoskeleton.
We train policies for 10 training human subjects and test the learned policies for 5 new human subjects.
Figure \ref{fig:SR_test} compares the average performance of EAP with DR and UP.
The larger area of stability region indicates that EAP significantly outperforms two baselines.

\subsection{Ablation study}
We further analyze the performance of EAPs by conducting a set of ablation studies.
We studied four categories of parameters: choices of observable parameters, reference dynamics, error prediction horizons, and error representations.

\begin{figure}
\centering
\setlength{\tabcolsep}{1pt}
\renewcommand{\arraystretch}{0.7}
\includegraphics[width=0.4\textwidth,height=5cm]{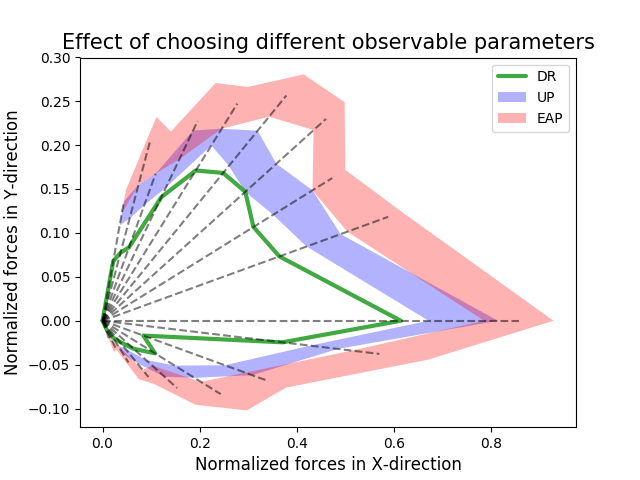}
\caption{\small{Ablation study with choosing different observable parameters as $\bm\mu$. The result indicates that our approach (EAP) shows more reliable zero-shot transfers for all different scenarios.}}
\label{fig:AblationMuo}
\end{figure}

\noindent\textbf{Choice of Observable and Unobservable Parameters.} We check the robustness of EAPs by testing with different choices of observable and unobservable parameters. We randomly split the parameters into $\bm{\mu}$ and $\bm{\nu}$ and test three different splits. %$\bm\mu$. \revised{To this end, we change the observability of some parameters, such as making joint damping unobservable or making PD gains as observable. We test three configurations of $\bm\mu$ and $\bm\nu$.}
% \sehoon{What about $\bm\nu$? The reviewer may ask it.} We test three different $\bm\mu$ for the assistive walking environment, which are ???, ???, and ???. \sehoon{fill the details.}
Figure \ref{fig:AblationMuo} shows the stability regions for all three algorithms for three different scenarios.
% , where the shaded region indicates the variation in performance for three different sets of $\bm\mu$.
% \karen{The choice of $\bm{\mu}$ is a discrete change, but the results is visualized as continuous shaded region. That seems a bit strange to me. How did you even compute the shaded region? And how come DR doesn't have any shaded region? Do all three choices for DR result in identical stability region?}.
In all cases, EAPs are more robust than the baseline algorithms.

\begin{figure}
\centering
\setlength{\tabcolsep}{1pt}
\renewcommand{\arraystretch}{0.7}
 
  \includegraphics[width=0.4\textwidth,height=5cm]{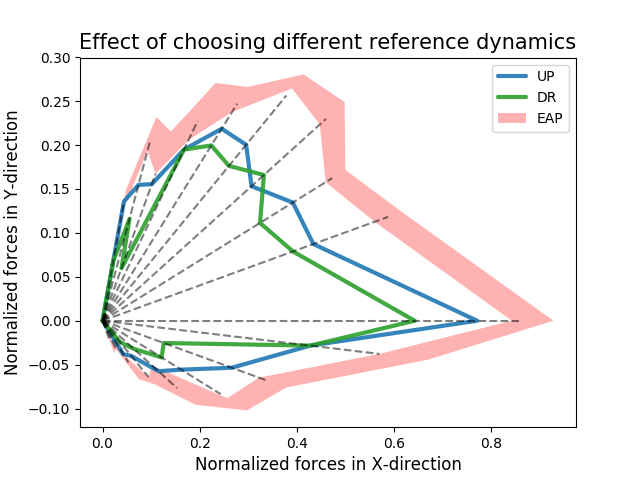}

\caption{\small{Ablation study with different reference dynamics. The results indicate that our algorithm is robust against the choice of different references.}}
\label{fig:ablationRefDyn}
\end{figure}

\noindent\textbf{Choice of Reference Dynamics.} In this study, we analyze the effect of choosing three different reference dynamics $P(\vc{s}' | \vc{a}, \vc{s}, \bm{\mu}_0, \bm{\nu}_0)$ on the performance of EAP. We randomly choose three different human agents as the reference dynamics and follow the learning procedure of EAPs to train three different policies. These policies are then deployed on the same test subjects along with UP and DR policies. Figure \ref{fig:ablationRefDyn} shows that all the EAPs outperforming the baselines by having larger stability regions, although EAPs have slightly larger variances.

\begin{figure}
\centering
\setlength{\tabcolsep}{1pt}
\renewcommand{\arraystretch}{0.7}
\includegraphics[width=0.4\textwidth,height=5cm]{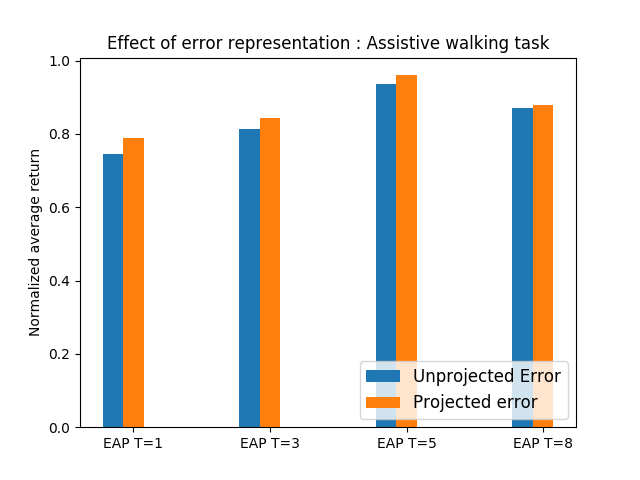}
\caption{\small{Ablation study with different parameter setting for EAP training.}}
\label{fig:ablation}
\end{figure}

\noindent\textbf{Horizon of Error Prediction.} 
As we motivated in Section~\ref{sec:error_func}, one step error might be too subtle to inform the learning of EAPs and we may need $T$ step expansion to enlarge them.
We studied the effect of the error prediction horizon $T$ in Algorithm~\ref{alg:errorfunc} by varying its value from $T=1$ to $T=8$ for the assistive walking task.
Figure \ref{fig:ablation} shows the normalized average return over T gradually changes over the different values of $T$ and peaks at $T=5$. 
Therefore, we set $T=5$ for all the experiments.

\noindent\textbf{The error representation.} 
We also compare the effect of the error representation.
Figure \ref{fig:ablation} also plots the normalized average returns of the unprojected errors (blue) and projected errors (orange), where projected errors show slightly better performance for all the different $T$ values.

\section{Conclusions}
We presented a novel approach to train an error-aware policy (EAP) that transfers effectively to unseen target environments in a zero-shot manner. Our method learns an EAP for an assistive wearable device to help a human recover balance after an external push is applied. We show that a single trained EAP is able to assist different human agents with unseen biomechanical characteristics. We also validate our approach by comparing EAP to common baselines like Universal Policy and Domain randomization to show our hypothesis that a policy which explicitly takes future state error as input can enable better decision making. Our approach outperforms the baselines in all the tasks. We also evaluated the performance of our algorithm through a series of ablation studies that sheds some light on the importance of parameters such as error horizon length, error representation, choice of observable parameters and choice of reference dynamics. We find that EAP is not sensitive to either the choice of observable parameters or the reference dynamics, and outperforms the baselines with variations in these quantities as well. 

Our work has a few limitations. At the core, our algorithm relies on the error function to make predictions of the expected state errors. The accuracy of this prediction can be improved by better function approximators such as recurrent neural networks (RNN) that takes a history of states as input, we leave this for future work. We also aim to test our approach on real-world robot.

\section*{Acknowledgments}

%% Use plainnat to work nicely with natbib. 

\bibliographystyle{plainnat}
\bibliography{References}

\end{document}